\def\year{2022}\relax
\documentclass[letterpaper]{article} 
\usepackage{aaai22}  
\usepackage{times}  
\usepackage{helvet}  
\usepackage{courier}  
\usepackage[hyphens]{url}  
\usepackage{graphicx} 
\usepackage{natbib}  
\usepackage{caption} 
\DeclareCaptionStyle{ruled}{labelfont=normalfont,labelsep=colon,strut=off} 
\usepackage{algorithm}
\usepackage{algorithmic}

\usepackage{subfig}
\usepackage{booktabs} 
\usepackage{enumitem}

\usepackage{newfloat}
\usepackage{listings}
\floatstyle{ruled}
\newfloat{listing}{tb}{lst}{}
\floatname{listing}{Listing}
\copyrighttext{Presented at the AI-HRI Symposium at AAAI Fall Symposium Series (FSS) 2022}

\title{Towards an Interpretable Hierarchical Agent Framework using Semantic Goals}

\author{
    Bharat Prakash\textsuperscript{\rm 1},
    Nicholas Waytowich\textsuperscript{\rm 2},
    Tim Oates\textsuperscript{\rm 1},
    Tinoosh Mohsenin\textsuperscript{\rm 1}, 
    \\
}
\affiliations{
    \textsuperscript{\rm 1}University of Maryland, Baltimore County,
    \textsuperscript{\rm 2}US Army Research Lab\\
}

\usepackage{bibentry}

\begin{document}

\maketitle

\begin{abstract}

Learning to solve long horizon temporally extended tasks with reinforcement learning has been a challenge for several years now.  We believe that it is important to leverage both the hierarchical structure of complex tasks and to use expert supervision whenever possible to solve such tasks. This work introduces an interpretable hierarchical agent framework by combining planning and semantic goal directed reinforcement learning. We assume access to certain spatial and haptic predicates and construct a simple and powerful semantic goal space. These semantic goal representations are more interpretable, making expert supervision and intervention easier. They also eliminate the need to write complex, dense reward functions thereby reducing human engineering effort. We evaluate our framework on a robotic block manipulation task and show that it performs better than other methods, including both sparse and dense reward functions. We also suggest some next steps and discuss how this framework makes interaction and collaboration with humans easier.

\end{abstract}

\section{Introduction}

Deep reinforcement learning has been successful in many tasks, including robotic control, games, energy management, etc. \cite{mnih2015human, schulman2017proximal, Warnell2018}. However, it has many challenges, such as exploration under sparse rewards, generalization, safety, etc. This makes it difficult to learn good policies in a sample efficient way. Popular ways to tackle these problems include using expert feedback \cite{christiano2017deep, Warnell2018, aaai2020-bharat} and leveraging the hierarchical structure of complex tasks. There is a long list of prior work which learns hierarchical policies to break down tasks into smaller sub-tasks \cite{sutton1999between, fruit2017exploration, bacon2017option, prakash2021interactive}. Some of them discover options or sub-tasks in an unsupervised way. On the other hand, using some form of supervision, either by providing details about the sub-tasks, intermediate rewards or high-level guidance is a recent approach \cite{prakash2021interactive} \cite{jiang2019language} \cite{le2018hierarchical}. 


\begin{figure}
        \centering{\includegraphics[width=0.47\textwidth]{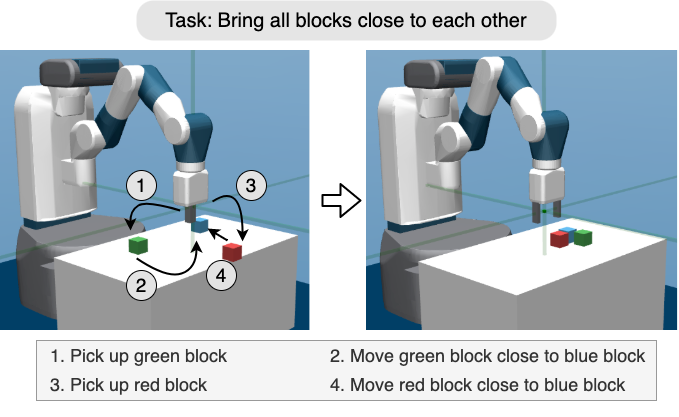}}
		\caption{Hierarchical agent: The figure shows a scenario where all the 3 blocks are placed far away from each other. The task is to bring all the blocks close to each other. This can be broken down into four subtasks as shown in the bottom. After the low-level policy of the agent executes all four subtasks, we get the desired final configuration where all three block are close to each other. }
		\label{fig:intro}
		\vspace*{-0.3cm}
\end{figure}

This paper presents a framework for solving long-horizon temporally extended tasks with a hierarchical agent framework using semantic goal representations. The agent has two levels of control and the ability to easily incorporate expert supervision and intervention. The high-level policy is a symbolic planner \cite{alkhazraji-et-al-zenodo2020} which outputs a plan in terms of sub-goals or macro-actions given an initial state and final goal state. The low-level policy is a goal-conditioned multi-task policy which is able to achieve sub-goals where these goals are specified using a semantic goal representation. The semantic goal representation is constructed using several predicate functions which define the behavior space of the agent. This representation has many benefits because it is much simpler than traditional state-based goal spaces as shown in \cite{akakzia2020grounding}. It is also more interpretable and easier for an expert to intervene and provide high-level feedback. For instance, given a high level goal, the planner finds a plan which can be observed by the human expert. It is easy to make small changes to the plan by adding sub-goals and changing the sequence if necessary. This framework also enables possibilities for collaboration. Due to the interpretability and ease of modifying the high-level plan, sub-tasks can be divided among agents and humans. This is not possible in other hierarchical agent frameworks where the high-level planner is also a black-box. 


We evaluate the framework using a robotic block manipulation environment. Our experiments show that this approach is able to solve different tasks by combining grasping, pushing and stacking blocks. Our contributions can be summarised as follows:
\begin{itemize}[noitemsep, topsep=0pt]
  \item A hierarchical agent framework where the high-level policy is a symbolic planner and the low-level policy is learned using semantic goal representations.
  \item Evaluation on complex long horizon robotic block manipulation tasks to show feasibility and sample efficiency
  \item A discussion showing the benefits of this framework in terms of interpretability and ability to interact and collaborate with humans.
\end{itemize}

\section{Methods}

In this section, we present a framework for solving long horizon temporally extended tasks. We first describe the problem statement with the environment used in our experiments. Then we describe the semantic goal representation and low-level policy training. Finally, we show how the high-level policy is obtained using the STRIPS planner \cite{FIKES1971189, alkhazraji-et-al-zenodo2020} to tie everything together and solve long horizon tasks.

\begin{figure}[ht]
        \centering{\includegraphics[width=0.42\textwidth]{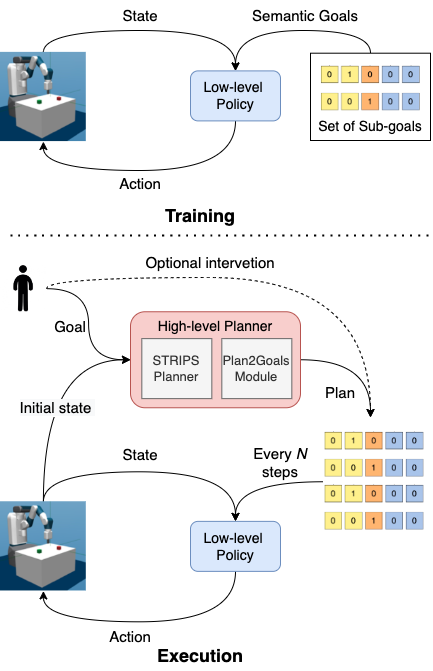}}
		\caption{The low-level policy receives semantic goals and is trained to execute primitive actions in the environment to achieve the goal. The set of sub-goals is curated by a human and can also involve a curriculum. To perform high-level tasks during execution, a human can specify a goal and the high-level planner outputs a plan, a sequence a sub-goals. A human can observe this plan and intervene/modify if required. The low-level policy achieves these sub-goals sequentially to perform the task.}
		\label{fig:hrl_arch}
\end{figure}

\subsection{Problem statement}

We aim to learn robotic control tasks in the Fetch Manipulation environment built on top of Mujoco \cite{todorov2012mujoco} which consists of a robotic arm with a gripper and square blocks. The observation space consists of the arm state including positions and velocities, the gripper state, and the Cartesian positions of the blocks. The robot can pick up, push and move the blocks. We built several tasks in this domain. We have the ability to initialize the scene with different configurations, like the block positions and robot arm and gripper positions.

\begin{figure}[t]
        \centering{\includegraphics[width=0.4\textwidth]{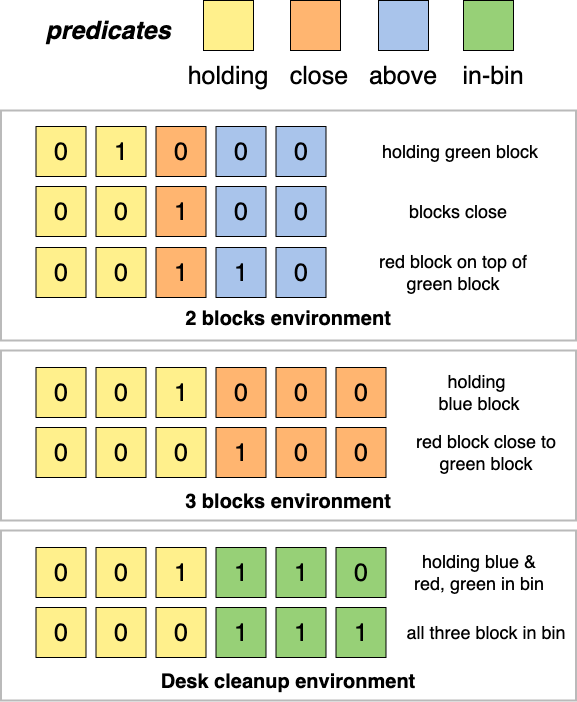}}
		\caption{Semantic Predicates: This figure shows how the semantic goal representations are created using the predicates. We consider three predicates \emph{holding, close, above} as shown. All three predicates are used in the 2 blocks environment. In the 3 blocks environment, we only use the \emph{holding} and \emph{close} predicates to simplify the problem. In the desk cleanup environment, we use the  \emph{holding} and \emph{in-bin} predicates. The above figure shows examples of a few configurations.}
		\label{fig:pred}
\end{figure}
\vspace*{-0.3cm}

\subsection{Semantic goal representations}

We represent goals using a list of semantic predicates which are determined based on domain knowledge. In our case we consider three spatial predicates - \emph{close}, \emph{above}, \emph{in-bin} and one haptic predicate - \emph{holding}. As demonstrated by \cite{akakzia2020grounding}, these predicates define a much simpler behavior space instead of the traditional more complicated state space. This makes it easier to represent goals and also define a curriculum as we will discuss in the next section. And more importantly, this representation eliminates the need to write reward functions for every desired behavior. 

All these predicates are binary functions applied to pairs of objects. The \emph{close} predicate is order-invariant. $close(o_1, o_2)$ denotes whether objects (in our case blocks) $o_1$ and $o_2$ are close to each other or not. The \emph{above} predicate is applied to all permutations of objects. $above(o_1, o_2)$ is use to denote if $o_1$ is above $o_2$. The \emph{in-bin} predicate is used to denote whether the block is inside the bin. Finally, \emph{holding} is used to denote if the robot arm is holding an object using $holding(o)$. With these predicates we can form a semantic representation of the state by simply concatenating all the predicate outputs as  shown in Fig \ref{fig:pred}.

\subsection{Training the low-level policy}

The low-level policy is trained to perform several individual sub-tasks, which can eventually by used to solve longer high-level tasks.
We use Hindsight experience replay (HER) \cite{andrychowicz2017hindsight} along with Soft-Actor critic (SAC) \cite{haarnoja2018soft} to train the goal conditioned policy. Goals are sampled from a set of configurations based on the environment where an expert can be used to optionally create a curriculum. The semantic goal representation makes is easier to do both of these things. The agent explores the environment to collect experience and updates its policy using SAC. As stated earlier, there is no need to write reward functions for each desired behavior. A reward can be generated by checking whether the current semantic configuration matches the goal configuration. The sub-goals for the two environments we use are listed in Table \ref{goals}. 

\begin{table}[h]
\small
\centering
\begin{tabular}{@{}lll@{}}
\toprule
\textbf{2 Blocks} & \textbf{3 Blocks}  & \textbf{Desk Cleanup} \\ \midrule
Pick X        & Pick X     & Pick X   \\
Put X near Y        & Put X near Y    & Put X on Table    \\
Put X away from Y        &  Put X away from Y  & Put X in Bin  \\
Put X on top of Y        &     &    \\ \bottomrule
\end{tabular}
\caption{Low-level policy is trained on the above set of semantic goals. The semantic goal representation is built using the predicates as described in the previous section. For 2 blocks version, X/Y can be block of color [\emph{red, green}]. For the 3 blocks version it can be of color [\emph{red, green, blue}]} 
  \label{goals}
\end{table}

\subsection{High-level planner}

\begin{figure*}[t]
        \centering{\includegraphics[width=1\textwidth]{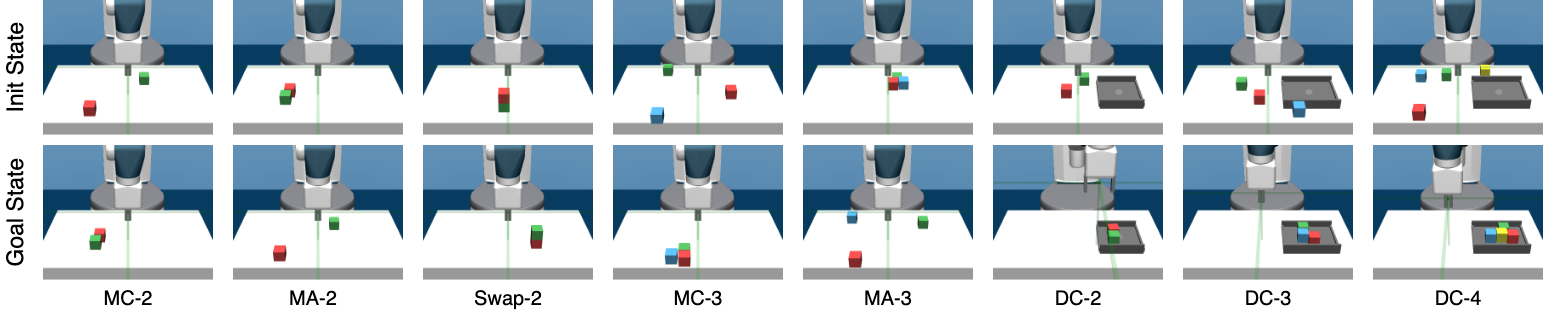}}
		\caption{This figure shows all the tasks we used in our experiments. The top row shows examples of random initial states and bottom row shows the goal states. MC-2: Move 2 blocks closer, MA-2: Move 2 blocks away, Swap-2: Swap 2 stacked blocks, MC-3: Move 3 blocks closer, MA-3: Move 3 blocks away, DC-2 to DC-4: Desk clean up with 2, 3 and 4 blocks}
		\label{fig:tasks}
\end{figure*}

We use a STRIPS planner \cite{FIKES1971189, alkhazraji-et-al-zenodo2020} as a high-level policy which provides sub-goals to solve the task. The STRIPS planner uses an encoding called Planning Domain Definition Language (PDDL) \cite{aeronautiques1998pddl} to represent the planing task. This planner can be defined using a 5 tuple (O, P, $S_i$, $S_g$, A). Here O represents the objects of interest, P is the predicates, $S_i$ and $S_g$ represent the initial and goal states respectively, and A represents the action space, in this case the high level macro-actions. 

The planner  accepts an initial state, $S_i$, and goal state, $S_g$, represented using objects and predicates. It then uses a search algorithm like BFS, A* etc. to find a plan to reach $S_g$ using actions A. The plan which is in PDDL is then translated to semantic goals which the low-level policy understands by $Plan2Goals$ module, a function which converts the PPDL output to predicates and semantic goals. We execute the low-level policy for a fixed number of steps $N$,  before switching the control back to the high-level sketch and execute the next sub-policy.

\begin{table*}[t]
  \centering
    \begin{tabular}{l c c c c c c c c }
    \toprule
      & \multicolumn{8}{c}{\textbf{Tasks}}\\ 
      \cmidrule(l){2-9} 
     \textbf{Method} & \textbf{MC-2}  & \textbf{MA-2} & \textbf{Swap-2} & \textbf{MC-3} & \textbf{MA-3} & \textbf{DC-2} & \textbf{DC-3} & \textbf{DC-4}\\
     \toprule
    Flat Semantic  &   10\%    &   80\% & 0\% & 5\% & 10\%  & 30\% & 0\% & 0\%  \\
    Flat Continuous &    5\%   &    10\% & 0\% & 0\% & 0\% & 0\% & 0\% & 0\% \\
    \textbf{Hierarchical Semantic (Ours)} &   \textbf{95\%}    &   \textbf{100\%}  & \textbf{92\%} & \textbf{95\%} & \textbf{96\%} & \textbf{94\%} & \textbf{91\%} & \textbf{90\%} \\
    \bottomrule
    \end{tabular}
    \caption{\textbf{Task completion \%} This table shows the task completion \% for our experiments. The tasks names are explained in Figure 4. As seen, our method consistently outperforms all the other baselines. Our method is the only one which can consistently solve all the three tasks. We train each agent for 2M steps and roll out 50 episodes using the trained policy. The values are an average of runs from three different seeds. } 
  \label{results_tab}
\end{table*}%

\section{Experiments}


\subsection{Environment setup and tasks}

We design two versions of the Fetch manipulation environment with 2 and 3 blocks. \\
\textbf{2 blocks environment} Here we have the robotic arm as mentioned earlier and two blocks: red and green. We consider all three predicates for this version, \emph{close}, \emph{above} and \emph{holding}. We design 3 high-level tasks in this environment (1) Move blocks close: Here the task is initialized with the 2 blocks far away from each other. The goal is to bring them close to each other. (2) Move blocks apart: Here the task is initialized with blocks close to each other. The goal is to move them apart. (3) Swap blocks: Here the task is initialized with blocks on top of each other in random order. The goal is to swap the order. \\
\textbf{3 blocks environment} Here we have the robotic arm and 3 blocks: red and green and blue. For this version, we only consider 2 predicates, \emph{close} and \emph{holding}. We design 2 high-level tasks in this environment (1) Move blocks close: Here the task is initialized with all the 3 blocks far away from each other. The goal is to bring them close to each other. (2) Move block apart: Here the task is initialized with blocks close to each other. The goal is to move them apart. \\
\textbf{Desk cleanup environment} Here we have a robotic arm and several blocks on the desk. The desk also a bin and the blocks are places randomly on the desk. The task is to clean up the desk and place all the blocks inside the bin. We use 2 predicates here, \emph{holding} and \emph{in-bin}. We have 3 versions with 2, 3 and 4 blocks.

\subsection{Baselines}

\emph{1. Flat semantic}: Here the agent has a single level policy but the goals are still represented using the semantic goal representations. There is no need to write reward functions for this version. \emph{2. Flat Continuous}: Here the goals are represented using the actual block positions of the desired configuration. The dense reward function is based on the distance between current and desired block locations and hence it is a dense reward function. 

\subsection{Results}

We calculate task completion \% for all the tasks using the fully trained agent. We train each agent for 2M steps and roll out 50 episodes using the trained policy. The values are an average of runs from three different seeds.

Table \ref{results_tab}  show the results for the 2 blocks, the 3 blocks and the desk cleanup environment. All the models are trained for 2M steps. As shown in the table, our method is able to solve all the eight high-level tasks. All the other baselines struggle to solve tasks. This is consistent across all the tasks.  In the \emph{move blocks apart} task represented as MA-2, shown in table \ref{results_tab}, the flat semantic is able to learn the task but we noticed that it learns an aggressive policy where it knocks one of the blocks away from the table which not a desirable behavior. Whereas our model gently picks a block and moves is away from the other block. 

To summarize, all the other baselines with and without dense reward signals fail to learn a good policy. We also performed experiments where we let the policy run for 5M steps and the baselines were unable to solve the task. This shows that using semantic predicates in the low-level policy and symbolic planner as the high-level policy truly helps to solve complex long-horizon tasks.

\section{Discussion and Conclusion}

In this paper we show that combining a symbolic planner and a low-level goal conditioned reinforcement learning policy is indeed a promising approach to build hierarchical agents. As the high-level planner and low-level policy communicate using semantic predicates, the framework is very interpretable.
This also makes it easier for a human to intervene at the high-level to provided appropriate sub-goals in case the plan needs to be modified. If the environment allows multiple actors, this framework can also enable collaboration by dividing sub-tasks among them.  For instance, consider the desk cleanup task with 4 blocks scattered around the table. The planner outputs a plan with 8 sub-tasks to clean the desk. It is easy to detect independent sub-tasks - pickup red block and place in bin, pickup green block and place in bin etc. Such sequences can be assigned to a human or a second agent and the planner can re-plan which is inexpensive and fast.

There are several directions in which this framework can be extended. Currently, the human has to communicate the goals using the predicates which is already much easier than using raw states. But this could be improved by building a language interface which can translate natural language sentences to semantic goals and PDDL for the planner. With the current state space, we assumed access to predicate functions. But with more complex observation like images, one can learn these predicate functions using a small amount of labelled data. To further demonstrate the capabilities of the framework we plan to perform experiments on more complex environments, real robots and qualitative analysis using human subjects. This work is a step towards simple and interpretable hierarchical agents and we hope to build upon it.

\nocite{aidin2021aicas, shiri2022efficient, navardi2022toward, flairs19-bharat}

\section{Acknowledgments}

This project was sponsored by the U.S. Army Research Laboratory under Cooperative Agreement Number W911NF2120076.

\bibliography{rl.bib, eehpc.bib}

\end{document}